\documentclass[sigconf]{acmart} 

\AtBeginDocument{%
  \providecommand\BibTeX{{%
    \normalfont B\kern-0.5em{\scshape i\kern-0.25em b}\kern-0.8em\TeX}}}

\copyrightyear{2021} 
\acmYear{2021} 
\setcopyright{acmlicensed}\acmConference[AIES '21]{Proceedings of the 2021 AAAI/ACM Conference on AI, Ethics, and Society}{May 19--21, 2021}{Virtual Event, USA}
\acmBooktitle{Proceedings of the 2021 AAAI/ACM Conference on AI, Ethics, and Society (AIES '21), May 19--21, 2021, Virtual Event, USA}
\acmPrice{15.00}
\acmDOI{10.1145/3461702.3462631}
\acmISBN{978-1-4503-8473-5/21/05}

\begin{document}

\title{Towards Accountability in the Use of Artificial Intelligence for Public Administrations}

\author{Michele Loi}
\affiliation{%
   \institution{University of Zurich}
   \city{Zürich}
   \country{Switzerland}}
\email{michele.loi@uzh.ch}

\author{Matthias Spielkamp}
\affiliation{%
   \institution{AlgorithmWatch}
   \city{Berlin}
   \country{Germany}}
\email{spielkamp@algorithmwatch.org}


\begin{abstract}
We argue that the phenomena of distributed responsibility, induced acceptance, and acceptance through ignorance constitute instances of imperfect delegation when tasks are delegated to computationally-driven systems. Imperfect delegation challenges human accountability. We hold that both direct public accountability via public transparency and indirect public accountability via transparency to auditors in public organizations can be both instrumentally ethically valuable and required as a matter of deontology from the principle of democratic self-government. We analyze  the regulatory content of 16 guideline documents about the use of AI in the public sector, by mapping their requirements to those of our philosophical account of accountability, and conclude that while some guidelines refer to processes that amount to auditing, it seems that the debate would benefit from more clarity about the nature of the entitlement of auditors and the goals of auditing, also in order to develop ethically meaningful standards with respect to which different forms of auditing can be evaluated and compared.
\end{abstract}




\maketitle

\section{Introduction}

Most ethics or organizational guidelines about the use of Artificial Intelligence (AI) mention the value of accountability \cite{jobin_global_2019}. Unsurprisingly, accountability is also mentioned as a goal in some recently published guidelines concerning the use of  AI in the public sector we consider.\footnote{See section 8 for inclusion criteria. We include also the Alan Turing document \cite{leslie_understanding_2019}, because it is explicitly referred to as guidance in the UK Government guidelines \cite{government_digital_service_and_office_for_artificial_intelligence_uk_guide_2019}.}
As we shall see, there are diverse reasons for this. Accountability, as clarified below (section 3) includes the element of answerability, which appears to be challenged by automation, especially some computationally peculiar forms of it. 

Our main contribution to the debate on AI accountability is twofold: first, we consider non-instrumental arguments for accountability grounded in democratic theory; second, we distinguish also between direct public accountability via public transparency and indirect public accountability via transparency to auditors. We argue that both can realize public accountability. In addition to defending our conceptual framework, we illustrate its empirical fruitfulness by showing that some practical requirements in 16 guidelines on AI in the public administration address each of the main issues our theoretical analysis unpacks.

We define the scope of accountable process in terms of automation that is computationally-driven, i.e., automation that avails itself of algorithms implemented by computing machines. We do not limit our attention to AI, in some restricted meaning of it, e.g. as including only the most advanced forms of machine learning. Accountability challenges do not derive only from computational models that cannot be described in the form of rules programmers themselves understand — the so-called black box models \cite{de_laat_algorithmic_2018}, \cite{kroll_accountable_2016}. We doubt, first of all, that black box models and their lack of transparency are the only reason why accountability for AI deserves discussion. Second, we doubt — along others in the literature \cite{kroll_accountable_2016} — that black-box models, in spite of the depth of the transparency problem they raise, make accountability impossible or \textit{sui generis}. Both assumptions explain why the scope of our analysis is quite broad and not limited to so called black-box models.\footnote{For a recent analysis overview of black box and explainable AI models see:\cite{arrieta_explainable_2019},\cite{belle_principles_2020}.} A similar narrow view, which we do not accept, is that instances where decisions are "fully automated" are the only case why discussing algorithmic accountability is important. We reject this view for two reasons as well: first, it is not clear what it means for a decision to be fully automated, given that automation is always controlled by some human agent responsible for it; and second, even if a sound definition of the distinction were given, it would fail to correspond to salient ethical differences —  e.g., partial automation in the criminal justice domain, such as using a software to calculate risk scores, may be more deserving of attention than full automation in a different domain, e.g., fully automated translation of foreign company news on an English language financial newspaper.

Since we appeal to neither opacity in the sense of black-box models nor to full automation in the framing of our analysis, we owe the reader a distinct analysis of the problem AI poses for accountability. Thus, our paper starts by providing a theoretical account of what accountability and its value are, in general, and in relation to automation, before delivering the empirically informed part of the pa-per, which is based on the analysis of 16 guidelines. Thus, our approach is a combination of a philosophical account of accountability for computation-driven automation and an empirically informed, descriptive analysis of guideline recommendations. This paper combines the two approaches in a way that we hope our interdisciplinary readers will find to be both refreshingly new and particularly insightful.  

The analysis of the content of guidelines shows that they can be interpreted as addressing a general accountability problem, namely one resulting from technological delegation, as opposed to an accountability gap specifically due to features of recent AI techniques. Indeed, many non-computational systems and circumstances raise the same challenges to accountability that we explore in the context of AI, and thus the accountability issues we are exploring are not, necessarily, distinctive to AI. On the other hand, it seems that the debate would benefit from more clarity about the nature of the entitlement of auditors and the goals of auditing, also in order to develop ethically meaningful standards with respect to which different forms of auditing can be evaluated and compared. Thus, the distinctions introduced here can improve the clarity of the goals of advocating accountability for AI-based systems.

Let us then turn to an overview of the paper. As announced, we start (section 2) by analyzing automation as a delegation process and the possible challenges for human accountability it poses. Section 3 provides the conceptual analysis of accountability that will be used in the rest of the paper. Section 4, 5, and 6 deal, respectively, with responsibility identification, public transparency, and auditing (or auditability). These three sections differ from the preceding two because they are not purely theoretical. Rather, we provide a synthesis of the recommendations included in 16 guidelines about the use of AI or algorithms in the public sector that are relevant to promoting account-ability according to our definition of it. We wrap up the paper with the conclusion, summarizing our main findings.

\section{AI and Automation}
By automation of decision-making, we mean the delegation of a subordinated cognitive or decisional function from an agent capable of accountability, i.e. a human,\footnote{As long as silicon-based intelligence will not have the features necessary for human-level agency.} to a non-biological form of information processing that has been designed by a human by specifying specific rules of computation. 
\begin{figure}[ht]
\centering
{\includegraphics[width=1\linewidth]{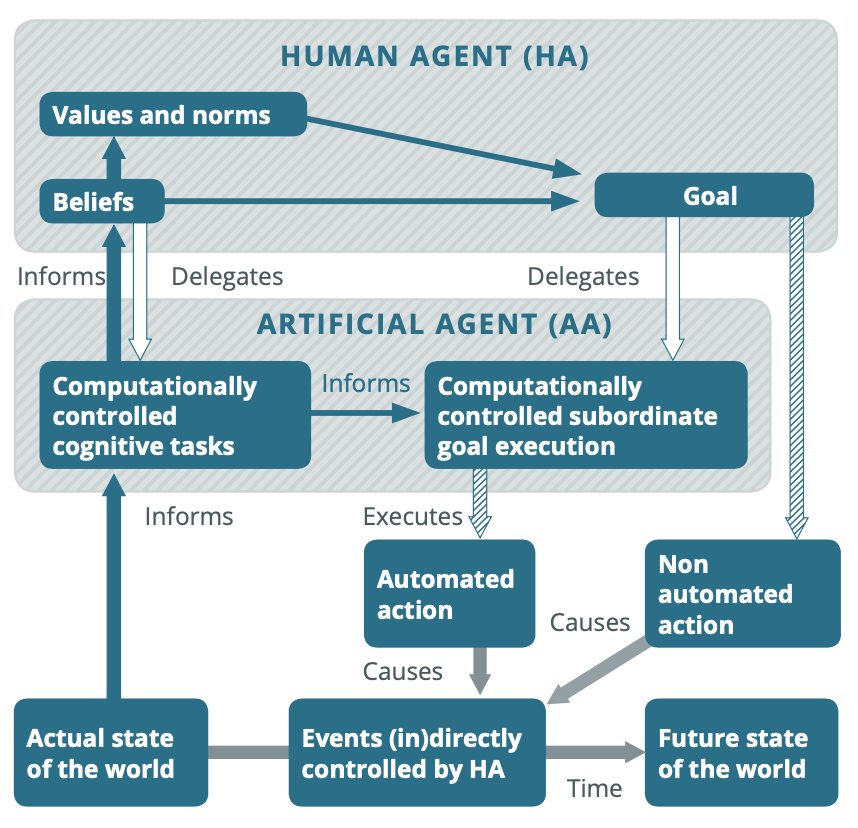}
\label{fig:delegation}}
\hfill
\caption{This figure represents an idealized automation process of both cognitive and executive functions. Full arrows denote information flow. White arrows denote delegation. Grey arrows denote causation. Oblique striped arrows denote execution (controlled causation).}
\end{figure}

The human agent (HA) can delegate either cognitive tasks or the execution of subordinated tasks, or both, to an artificial agent (AA). The delegation is ideal if and only if all subordinate tasks and cognitive tasks adequately contribute to HA's goal as intended. HA indirectly controls the out-comes in spite of the automation of subordinated tasks, by dictating the overarching goals, which control all the most important parameters or boundary conditions for actions by AA. Hence, HA controls 1) own actions directly; 2) actions of AA indirectly, in so far as a reasonable guarantee exists that they merely implement HA's will; 3), HA has perfect "higher-order" cognition of her relation to AA, i.e., full awareness that the delegation of some cognitive tasks to the AA has a feedback on the HA's own beliefs and, potentially, value system. 

As we shall see, the above given picture of delegation describes an idealized condition of decisional autonomy for HA. In these circumstances, HA retains full moral and causal responsibility for HA's actions as well as AA's actions. However, such idealized picture of human-machine interaction rarely occurs in the real world.

To the contrary, we witness three challenges to real accountability, namely distributed responsibility, externally induced automation acceptance, and automation acceptance through ignorance, the two latter conditions being forms of lack of meaningful control.
\paragraph{Distributed responsibility.} First of all, when the human agent belongs to a complex organizational structure, the nature of delegation may be hard to reconstruct. HA1 is a human resource employee and she uses an applicant rating software based on automated scans of applicants’ CVs (AA1). HA1 has received inadequate training about the AA1's limitations and is not aware she is using it for cases that are not suited to it. Management strongly encourages using AA1 as the general case. Not only does HA1 have little awareness of the software's limitation, but, given the extant company culture and time constraints of her role, HA1 feels she has little alternative to relying on the tool in every case.\footnote{Where the use of a system is mandated by an employer, this is included in both the category of ‘distributed responsibility’ and of ‘externally induced automation acceptance’. The case discussed here is an instance of both and it is necessarily a failure of meaningful control by virtue of being an instance of external inducement. But it is not a failure of meaningful control by virtue of distributed responsibilities, because responsibilities can be distributed in an egalitarian way. When this is the case, the use of technology is not necessarily externally induced and meaningful control may be preserved. Still, distributed responsibility makes it difficult to determine who should be held accountable if a problem persist, as we shall see in section 3.} If management choices had been different (e.g., training, incentives, time constraints) she would not be using the tool all the time and would spend more time evaluating the candidates independently. This points to responsibilities of the management,  the fact that this AA1 does not serve the interest of a "single boss" (her direct user, HA1), but interacts within a system  of distributed responsibility \cite{floridi_faultless_2016}, thus raising the (accountability) "problem of many hands"\cite{wieringa_what_2020}. 
\paragraph{Lack of meaningful control.} Even though HA1 treats the artificial agent as an adequate means to her goals, AA1's behavior (e.g., choices) does not track what HA1 believes are good reasons. In the case of cognitive delegation, the good reasons in question are epistemic, e.g., reasons for making specific inferences. In the case of executive delegation, reasons are practical, i.e., reasons to make choices\cite{santoni_de_sio_meaningful_2018}. We shall distinguish two preventative conditions of meaningful control:\textit{induced acceptance} and \textit{ignorant acceptance}.
\paragraph{Externally induced automation acceptance.}Whereas HA1, AA1's user, relies on AA1 as an adequate means to her own goals, tracking her own reasons, another agent, not a user, has imposed goals and requirements that are incompatible with or strongly suboptimal to achieve AA's over-arching goals, or lead to significant undesired collateral effects from AA's point of view. HA2's friends all use the same social network app. To fulfill her sociality needs, HA2 also uses the same social network app. The algorithm of this app, AA2, influences her beliefs, nudges her actions, and makes autonomous decisions (e.g., with respect to what content to prioritize on the medium's feed). But the social network is designed to maximize the time HA2 spends staring at her device's screen. That takes place at the expense of alternative socialization activities, such as HA2's spending time outside with her friends, that would actually be more rewarding from HA2's viewpoint. This can happen because the design (e.g., of nudges) is optimized for a goal different from her own and it is too costly to avoid relying on the AA. HA2 does not meaningfully control AA2 because AA2's goals are not sufficiently aligned with HA2's. HA2 accepts AA2's goal only because they are bounded with a form of automation she has most reasons overall to accept, given the lack of equally desirable alternatives.
\paragraph{Automation acceptance through ignorance.} Often, the human agent is not in the position to understand the capabilities of the system and the way in which it takes decisions \cite{santoni_de_sio_meaningful_2018}. HA3 uses an online dating app to find his romantic partner. HA3's view of the ideal partner is however so misguided that it makes him unlikely to achieve romantic success. HA3 consented to a randomized experiment intended to test the efficacy of the matching algorithm. By ending up in the control arm of the experiment, HA3 is assigned with the poorest possible match according to the app's own algorithm. In spite of signing a consent form, HA3 does not understand this. The app finds HA3's for the first time in his life a matching partner and this happens precisely because, unbeknownst to HA3, he finally gave an opportunity to someone who contradicted all of HA3’s desiderata. HA3 lacks meaningful human control even if the app does what is in HA3’s ultimate interest. HA3 does not control the app he relies on in any meaningful sense, because he does not understand enough of what the app does and why, witness the fact that he would have bounced the partner proposed to him if he had understood how it came to be. This category includes human non-cognitive factors that explain why an individual uses the technology in those situations in which the individual would not use the technology if she were aware of them, as in e.g. automation complacency, automation bias, etc. 

The three phenomena, \textit{distributed responsibility}, \textit{induced acceptance}, and \textit{acceptance through ignorance} are pervasive of many people's relationships to automation. These are all instances of imperfect delegation. As we illustrate after having analyzed the concept of accountability, imperfect delegation threatens one or more key elements of accountability. Thus — our thesis goes — imperfect delegation challenges human accountability when tasks are delegated to computationally-driven systems. We are not rushing to the conclusion that HAs delegating functions to AAs is not morally responsible or accountable at all for their actions. After all, the action of delegation to AAs remains each HA's own action. But the implications for responsibility are clearly more complex than those described in ideal delegation sketched at the start.

We are not in the position to specify how we understand the expression "AI" which we used in the title of this contribution and the topic of "automated decision-making" that is an alternative often preferred to AI in some recent ethics/governance guidelines covering roughly the same (or at least an overlapping) terrain. What we (stipulatively) mean here by "automation of human decision-making" through "AI" is "any delegation of decision-making to computationally-driven systems with the potential to cause an accountability gap because of the three above highlighted phenomena".

It follows that our analysis does not focus on "black-box" models, but can explain why automation relying on black-box models can create such challenges (hence belongs to "AI" in our stipulated sense). The opacity of those systems arguably makes it difficult for all users to achieve meaningful human control, which arguably makes acceptance through ignorance more likely. Yet, explainability is relative to the cognitive abilities of the user \cite{ibnouhsein_quelle_2018}, so acceptance through ignorance is a more widespread phenomenon. In all three cases, the accountability of the user is compromised by imperfect delegation. Our thesis is that imperfect delegation leads to inadequate accountability of all the relevant human agents involved in the decision that have significant responsibilities in causing the relevant actions and effects. Such inadequacy, we argue, is especially problematic when the end user of automation acts in the name of the public administration. For in this specific case, the end user is morally and politically supposed to be accountable to the citizens.
\section{What is accountability?}
Accountability is a relational condition: it cannot be defined as the quality of an agent in isolation from other individuals. Any definition of accountability will include at least three elements:
\begin{enumerate}
    \item\textit{responsibility}, for actions and choices, which also provides the ground for moral praise or blame, social approval, and being liable to legal sanctions;
    \item\textit{answerability}, which includes two aspects:
    
    \begin{enumerate}
    \item\textit{capacity and willingness}, to reveal the reasons behind decisions to a selected counterpart (which may also be the community as a whole),
    \item\textit{entitlement} of such counterpart to request that such reasons are revealed; and finally (and somewhat less unanimously);
    \end{enumerate}
   \item\textit{sanctionability} of the accountable party \cite{boos_getting_2020},\cite{wieringa_what_2020}.
\end{enumerate}
Notice that C appears unduly restrictive (compared to most actual uses of accountability, especially in AI ethics discourse), unless "sanction" is understood in the broadest possible sense, which includes receiving moral blame, avoidance by other parties of commercial interactions, punishment by consumers, etc. and not just narrowly to mean punishment on the basis of law. 

What is the link between the three elements mentioned above and the three challenges mentioned above, i.e., distributed responsibility, induced acceptance, and ignorance-driven acceptance? 

First of all, the problem of distributed responsibility poses a challenge to the identification of responsibility. If HA1's choice to delegate the evaluation of candidates is induced by company culture and time constraints, at least morally speaking HA1 is not the only person responsible for the resulting delegated human resources decisions. This also challenges sanctionability, because it is not obvious (morally at least) who should be sanctioned if the use of the software to make such human resource decisions results in, for example, unfairness.\footnote{For example, when a large organization is involved in a disaster, it is often difficult to obtain convictions for the most significant criminal charges in the courts. This is also due to flow of information that needs to be provided to, for example, people in charge for the design of a technology, about its harmful effects, in order to consider these individuals morally and legally responsible for the flaws the technology produces.}

Second, the problems of induced acceptance, and acceptance through ignorance arguably threaten the answerability dimension of accountability. Suppose the human resource user of AA1, HA1, is a public servant responsible for hiring in the public sector. HA1 provides truthful explanations of her reason: namely, she needs to rank candidates and in the context of her time requirements and education, (it looks as if) using the rating provided by the software is the best she can do to achieve her goal fairly and accurately. Instead, because of her poor training, HA1 ignores that she should not be using that software to make that particular decision for candidates for that specific position. The software is not robust and accurate in that type of use. Moreover, HA1 ignores why the software appears to be making the kind of ranking it does. Because of her ignorance, HA1 does not have meaningful human control of the task he delegates to the software. She does not understand the technology and its limitations well enough to employ either teleological reasoning \cite{loi_transparency_2020}, or causal/counterfactual reasoning \cite{wachter_counterfactual_2017} in providing an explanation or a (teleological) justification of the decision resulting from delegation. If so, even if HA1 truthfully reveals (what she takes to be) her reasons for an action, the reasons should not be considered satisfactory by any reasonable counterpart. Accountability should not be considered achieved in this case. (It is not achieved, either, by blaming HA1 for her poor judgment, or by sanctioning her for the resulting unfairness.)

Third, the entitlement dimension of accountability is compromised if, on the one hand, the public is only entitled to answers by individuals in the public administration, in particular end users, but not other parties with equally important responsibilities, who remain not accountable. In the public administration HR case, the employee whose promotion was rejected may be entitled to an explanation by the HR department, which may try to explain how the algorithm decided. But no explanation is due to an employee by HA1's boss. Also in the case in which the public administration uses a software that — by analogy to HA2's case — prioritizes other goals of the software designer, if the public is entitled to an answer by the public administrator, but not by the technology developer, answerability is obtained only formally, but not substantially.

The answerability challenge is particularly important for non-instrumental democratic theory \cite{boos_getting_2020}. This theory considers accountability of the government towards the governed as an essential element of democratic self-government. In the case of HA3, ignorance-driven acceptance occurred even though the app used decided in HA3's best interest. Consider now a case in which company A provides the public administration with software influencing high-stake decisions about members of the public. The public administration is as ignorant about the deep underlying logic of the software as HA3 is of the randomized experiment in which he is involved. A's CEO, however, is a more sensitive social thinker than anyone in the current government, and the principles she requires her software to implement are ethically and economically sounder than those the administration has asked the company to implement. As a result, the community is better off with decisions taken by A's software than it would have been if the software had only followed the specifications of the public administration.

The case at hand is analogous, for non-instrumental democratic theory, to that of a non-democratically accountable government that happens to promote the welfare of the population better than a democratic government would have. Irrespective of the good outcomes such government achieves, it is not a case of self-government. The same is true of the decisions of the public administration systematically influenced or based by an "AI" which is designed to achieve some goals or respect requirements imposed by a (non-publicly accountable) technology developer. In the best-case scenario in which CEO's of technological companies providing the public administration with software are reliable better than democratically appointed officials, if so much influence is permitted to obtain on public administration decisions, we are no longer dealing with democratic self-government but with a (benign) form of technocracy. 

\begin{figure}[h]
\centering
{\includegraphics[width=1\linewidth]{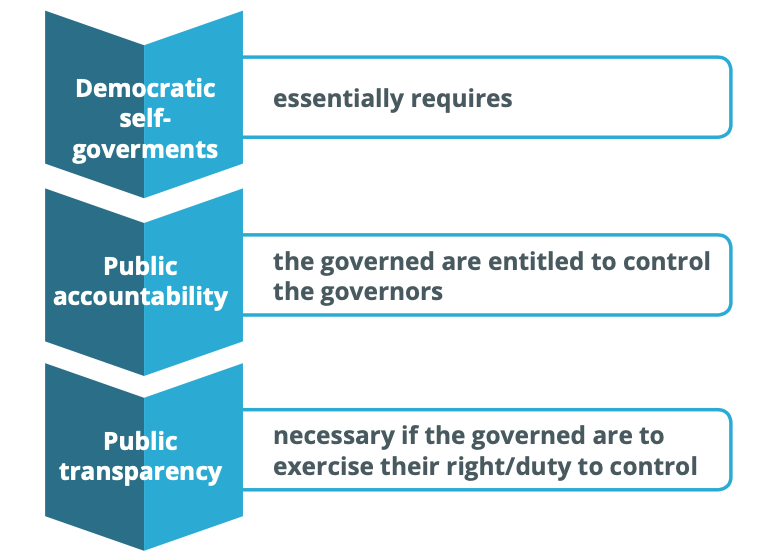}
\label{fig:delegation}}
\hfill
\caption{The relationship between accountability and public transparency, according to non-instrumental democratic theory.}
\end{figure}

In what follows, we take a closer look at some recently published guidelines on AI in the public sector to illustrate how they address the three distinct challenges to accountability that automation raises. We shall also often refer to other ethical values and principles, in particular those of beneficence, non-maleficence, autonomy and justice. The choice of these principles is dictated by two considerations: they are in widespread use in applied ethics, particularly bioethics \cite{beauchamp_principles_2008} and they are often invoked (entirely or selectively, alone or in conjunction with others) in many different ethical frameworks that have been proposed for the ethics of AI \cite{floridi_unified_2019},  \cite{independent_high-level_expert_group_on_artificial_intelligence_set_up_by_the_european_commission_ethics_2019}, \cite{jobin_global_2019}.

\section{Accountability and responsibility}
In this section, we use the conceptual framework of accountability and its challenges in the context of automation as a lens to interpret a diverse set comprising 16 ethical guidelines. We cite the document as an in-text citation and use footnotes to indicate the name of the specific guideline principle (in some cases, document section) in which the concept appears.

Let us begin by considering distributed responsibility, also known as the "problem of many hands". Several recommendations address this concern. In AI ethics guidelines, it is normally assumed that only human persons can be accountable, (current) AIs cannot. In keeping with the above given analysis of accountability, we classify under the heading of accountability measures invoked to ensure that "people in charge" can be identified (forward-looking responsibility) \cite{dawson_artificial_2020}\footnote{Cf. "Accountability."} and that unethical behavior by responsible agents can be identified and sanctioned (backward looking responsibility) \cite{government_of_new_zealand_algorithm_2020}.\footnote{Cf. "Human oversight and accountability."} These two elements are arguably the core elements of accountability, those that can more clearly be distinguished from transparency and safety. The guidelines we have examined require, for example, that organizations should "establish a continuous chain of responsibility for all roles involved in the design and implementation lifecycle of the project" \cite{government_digital_service_and_office_for_artificial_intelligence_uk_guide_2019}.\footnote{Cf. "Accountability"; see also: \cite{engelmann_ki_2020}, \cite{schweizerische_eidgenossenschaft_-_der_bundesrat_leitlinien_2020}, cf. "Leitlinie 4."} This in turn requires clearly documented, monitored, controlled processes and outcomes \cite{government_digital_service_and_office_for_artificial_intelligence_uk_guide_2019}\footnote{Cf. "Accountability."}— we analyse the relation between documentation and responsibility in details in section 4.

The elements of answerability and (less clearly, sanctionability) are invoked, indirectly, by those guidelines that aim to enable the contestation or challenge of the decisions taken by partially or fully automated systems
\cite{ai_now_institute_using_nodate},
\cite{council_of_europe_recommendation_2020},\footnote{Cf. "Contestability."}
\cite{dawson_artificial_2020}.\footnote{Cf. "Contestability."}
In some cases the concept of due process is  used \cite{ai_now_institute_using_nodate}, which involves a strong form of answerability: institutions deploying the AI are responsible for collecting the feedback of the people affected by it and to implement the required remedial actions
\cite{ai_now_institute_using_nodate};\footnote{Cf. "Participatory Democracy, diversity and inclusion."} see also \cite{council_of_europe_recommendation_2020}\footnote{Cf. "Consultation and adequate oversight."}
and \cite{dawson_artificial_2020},\footnote{Cf. "Recourse."} i.e., compensation for the harm suffered. 
\section{Accountability and transparency}
It is fairly common that accountability and transparency principles or sections of different guidelines include the same, similar, or overlapping prescriptions. We explain this by showing that accountability requires (some kind of) transparency. Our analysis can easily diversify the two concepts of transparency: the first, control transparency is a way to make information accessible and to communicate for any purpose; the second, transparency-as-a-right, implies an entitlement of a counterpart outside the accountable organization to obtain that information. Both are claimed to enable a range of ethically valuable effects, such as the identification of harmful errors (ethical principle of non-maleficence, or do no harm), alignment with user preferences, generating higher satisfaction (ethical principle of beneficence). Some ethically desirable effects of transparency require transparency-as-a-right. Clearly, the individual consent to AI uses of personal data implies transparency-as-a-right, not just control transparency. Individual consent is a necessary condition of certain forms of human autonomy. 

This section is split in two sub-sections, corresponding to two operationally and morally distinct forms of transparency: internal control and public transparency. Both kinds of transparency are related to accountability, but they are related to it in different ways, i.e., by virtue of different elements. Transparency as internal control is necessarily a form of control transparency; public transparency is necessarily a form of transparency-as-a-right (of the public).

\paragraph{Transparency as internal control.} To begin with, internal control includes the activity of timely documenting processes and outcomes and the recording \cite{dataethical_thinkdotank_white_nodate},\footnote{Cf. "Traceability."}
testing \cite{council_of_europe_recommendation_2020} 
and monitoring \cite{council_of_europe_recommendation_2020} of the relevant events.\footnote{Cf. "Interaction of systems."}  
These activities together produce the information about the processes that can be made transparent. Second, control includes recommended practices of measuring, assessing, evaluating \cite{ai_now_institute_using_nodate},\footnote{Cf. "Key Elements Of A Public Agency Algorithmic Impact Assessment, \#1." See also \cite{schweizerische_eidgenossenschaft_-_der_bundesrat_leitlinien_2020}, cf. "Leitlinie 3."} 
\cite{council_of_europe_recommendation_2020},\footnote{Cf. "Ongoing review," "Evaluation of datasets and system externalities," "Testing on personal data."}
\cite{world_economic_forum_ai_2020},\footnote{Cf. "Data Quality."}
\cite{automated_decision_systems_task_force_new_2019},\footnote{Cf. "Explanation."}
\cite{automated_decision_systems_task_force_new_2019}.\footnote{Cf. "Impact determination."} 
It includes defining standards \cite{council_of_europe_recommendation_2020}\footnote{Cf. "Standards."} and policies. Transparency as internal control includes 
explicability \cite{dataethical_thinkdotank_white_nodate},\footnote{Cf. "Explainability."}
\cite{government_digital_service_and_office_for_artificial_intelligence_uk_guide_2019},
\cite{government_of_new_zealand_algorithm_2020},\footnote{Cf. "Transparency."}
\cite{automated_decision_systems_task_force_new_2019}.\footnote{Cf. "Explanation."}
Transparency also requires justification \cite{leslie_understanding_2019},\footnote{Cf. "Transparency."}
\cite{council_of_europe_recommendation_2020}\footnote{Cf. "Testing."} for design choices and, when unavoidable, its errors, biases and trade-offs with other moral goals.

Third, control includes those social activities necessary to ensure that one’s study of processes and outcomes is adequately complete and that it does not exclude relevant perspectives. This includes activities such as training
\cite{council_of_europe_recommendation_2020},\footnote{Cf. "Personnel management."} 
enhancing internal expertise \cite{ai_now_institute_using_nodate},\footnote{Cf. "Executive Summary."} \cite{council_of_europe_recommendation_2020},\footnote{Cf. "Independent research" and "Rights-promoting technology."}
and expert review\cite{ai_now_institute_using_nodate},\footnote{Cf. "Key Elements Of A Public Agency Algorithmic Impact Assessment, \#2."}
\cite{treasury_board_of_canada_secretariat_directive_2019},\footnote{Cf. "Appendix C."}
\cite{council_of_europe_recommendation_2020}.\footnote{Cf. "Consultation and adequate oversight" and "Expertise and oversight."}
Even diversity in the workforce \cite{council_of_europe_recommendation_2020}\footnote{Cf. "Principle of Equality and Security" and "Personnel management."}
and transparency as public debate \cite{council_of_europe_recommendation_2020}\footnote{Cf. "Public debate."}
can be advocated as a means to improving the accountable party's understanding of the implications of AI \cite{automated_decision_systems_task_force_new_2019}.\footnote{Cf. "Available information."}

Fourth, control includes risk-mitigation measures, such as building backups and contingency plans \cite{treasury_board_of_canada_secretariat_directive_2019},\footnote{Cf. "Appendix C."}  
making room for human intervention \cite{government_of_new_zealand_algorithm_2020}, \cite{council_of_europe_recommendation_2020},\footnote{Cf. "Principle ‘under user control’."}
predicting and preventing risks, prohibiting harmful or risky practices \cite{council_of_europe_recommendation_2020},\footnote{Cf. "Consultation and adequate oversight" and "Follow up."}
and correcting errors that are made \cite{council_of_europe_recommendation_2020}.\footnote{Cf. "Consultation and adequate oversight" and "Effective remedies."} 
These are all safety practices for which people "in charge" of AI implementation in the public administration can be held accountable. The importance assigned to risk assessment and management \cite{world_economic_forum_ai_2020},\footnote{Cf. "Key variables to consider in a risk assessment."}
\cite{council_of_europe_recommendation_2020},\footnote{Cf. "Human Rights Impact Assessment."}
\cite{government_of_new_zealand_algorithm_2020},\footnote{Cf. "Assessing likelihood and impact". Cf. "Human oversight and accountability," "Reliability, Security and Privacy."}
\cite{treasury_board_of_canada_secretariat_directive_2019},\footnote{Cf. "Algorithmic Impact Assessment."}
in the guidelines we have analyzed can hardly by overstated.

Fifth, and of special importance for the use of AI in the public sector, control includes ownership, knowledge, and effective control of some key infrastructure \cite{council_of_europe_recommendation_2020},\footnote{Cf. "Infrastructure" and "Interaction of systems."}  e.g., data assets and the machine learning algorithms to learn from them, that is essential for shaping, better knowing, and more tightly controlling the AI in use.  

Transparency via internal control is required by the accountability dimension of responsibility identification. First, internal control is necessary in order to identify who should be held responsible for normatively relevant outcomes. Second, internal control is necessary in order to identify what individuals should be held accountable (including, sanctioned or supported) for. 

This illustrates the relation between internal transparency and accountability. Let us now turn to public transparency.

\paragraph{Public transparency.} Public transparency is, we maintain, a dimension of transparency distinct from internal transparency. By public transparency we mean exclusively the production and communication of information to the broader public, or, in terms of democratic theory, "the governed." 

There are at least four main normative theories why transparency is instrumentally valuable 
\cite{de_laat_algorithmic_2018}, \cite{felzmann_transparency_2019}, \cite{loi_transparency_2020}, \cite{zarsky_transparent_2013}.
First, there is the view that "sunlight is the best disinfectant," to cite Justice Louise Brandeis, that is to say, the view that public transparency promotes accountability, which in turn prevents at least the worst unethical behavior from occurring. This justification links transparency with accountability, but it assigns a purely instrumental value to the latter, i.e., the prevention of unethical behavior (which can be also spelled out as behavior violating other moral principles, first of all the harm prevention and the justice principles).

Second, there is the view that public transparency contributes to the quality of the technology, because it enables the crowd-sourcing of expert opinion and the feedback by concerned citizens, which leads to better scrutiny of the technology, which makes it more trustworthy. This justification is more closely associated with the ethical principle of beneficence.

Third, there is the view that public transparency enables end users of a technology, or people who may be affected by it, to make an informed choice whether to use it. This justification is more closely associated with the ethical principle of autonomy. 

Notice that the principles of autonomy and beneficence in the second and third reason justify public transparency independently of accountability. The first justification, the idea that public transparency generates incentives for more ethical behavior, instead, refers to accountability directly. (In so far as it implies the existence of sanctions, at least of the reputational kind.) Thus, public transparency is instrumentally related to better outcomes and improved respect of the four ethical principles of beneficence, non-maleficence, justice and autonomy, both directly and indirectly. 

Notice that, according to the three views examined so far, public transparency is only valuable contingently when it induces more ethical behavior on the accountable parties or when it leads to ethical outcome improvement directly, e.g., via crowdsourcing. When the behavior of the accountable party cannot be improved through transparency mechanisms, public transparency has no instrumental value. Hence, it is quite legitimate to be skeptical of the instrumental value of public transparency if the instrumental justification is the only one available and the evidence that transparency generates better outcomes is hard to find.

Fourth, there is the view that public transparency enables public debate which is necessary for the democratic legitimacy of technological solutions. This is especially important when the implementation of technology is not value-neutral. This value of public transparency, in this picture, is a non-instrumental value from the viewpoint of democratic theory. Public transparency is non-instrumentally required by democratic self-government. It is valuable independently of its ethical outcomes, if one assumes that democratic self-government also is valuable as an end in itself. According to this value theory, accountability need not incentivize ethical behavior in order to be ethically required.

The distinction between the instrumental and non-instrumental value of accountability is important because instrumental views are more vulnerable to empirical sociological objections 
\cite{felzmann_transparency_2019}, 
\cite{zarsky_transparent_2013}. Public transparency may not have equally significant outcomes in all domains of application of AI. The incentive and ability to control of the broad public may be limited to a few cases that grab the attention of the media, so it may generate poor incentives for ethical behavior. The (non-accountability) related instrumental justifications for public transparency do not easily justify a broad scope for transparency, but only specific forms of it. E.g., with regards to the crowd-sourcing justification, subjecting the technology to the scrutiny of a restricted and selected group of experts may often be enough to make technology safe. The autonomy theory (all people need transparency to determine which technology is better "for them") does not take into account the limited ability of ordinary individuals to assess technology beyond its usability and pleasantness or (e.g., as it is often the case for the public administration) the fact that ordinary individuals are not presented with meaningful options to choose from \cite{zarsky_transparent_2013}.

\begin{figure}[ht]
\centering
{\includegraphics[width=1\linewidth]{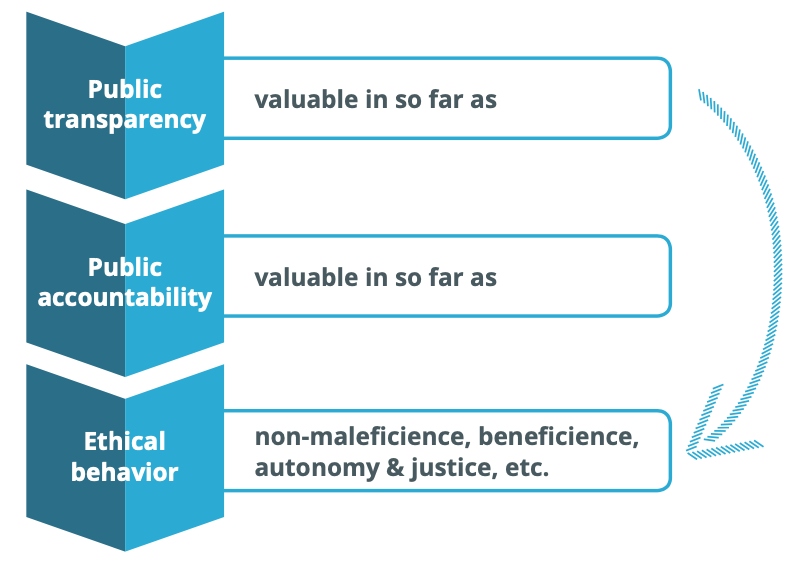}
\label{fig:delegation}}
\hfill
\caption{Relation between public transparency, accountability, and other ethical principles, assuming that accountability is only valuable instrumentally. Public transparency can facilitate ethical behavior directly or incentivize it indirectly (through accountability).}
\end{figure}
However, doubts about the contingent value of public transparency are irrelevant from the viewpoint of non-instrumental democratic theory. If that value theory is correct, public transparency is a deontological requirement whose value is independent of its effects. In other words: under conditions of democracy, citizens are entitled to hold public authorities accountable, independently of whether they take that opportunity and in this way generate outcome improvements. Answerability as a moral duty is a matter of deontological political ethics, not expediency. Yet, some may find the case for this deontological principle unpersuasive, especially if it turns out that, in practice, the public is either not interested, or not skilled enough, to participate in the relevant debates that public transparency is meant to enable.
\begin{table}[ht]
\small
\begin{tabular}{|l|l|l|}
\hline
\textbf{\begin{tabular}[c]{@{}l@{}}Function of \\ transparency\end{tabular}} & \textbf{Goals} & \textbf{\begin{tabular}[c]{@{}l@{}}Ethical principle\\  involved\end{tabular}} \\ \hline
\begin{tabular}[c]{@{}l@{}}Transparency \\ as disinfectant\end{tabular} & \begin{tabular}[c]{@{}l@{}}Accountability, \\ avoiding\\ unethical behavior\end{tabular} & \begin{tabular}[c]{@{}l@{}}Harm prevention, \\ beneficence, \\ autonomy, justice \\ (instrumental)\end{tabular} \\ \hline
\begin{tabular}[c]{@{}l@{}}Transparency for \\ crowd-sourcing\end{tabular} & \begin{tabular}[c]{@{}l@{}}Collecting expert\\  and lay people \\ opinion\end{tabular} & \begin{tabular}[c]{@{}l@{}}Beneficence\\ (instrumental)\end{tabular} \\ \hline
\begin{tabular}[c]{@{}l@{}}Transparency \\ for informed choice\end{tabular} & \begin{tabular}[c]{@{}l@{}}Enabling informed\\  individual choice\end{tabular} & \begin{tabular}[c]{@{}l@{}}Autonomy\\ (instrumental)\end{tabular} \\ \hline
\begin{tabular}[c]{@{}l@{}}Transparency \\ for informed \\ public debate\end{tabular} & \begin{tabular}[c]{@{}l@{}}Enabling informed\\ democratic deliberation\end{tabular} & \begin{tabular}[c]{@{}l@{}}Self-government\\ (deontological)\end{tabular} \\ \hline
\end{tabular}
\caption{\label{tab:table-name}Varieties of moral groundings for public transparency.}
\end{table}

Public transparency is invoked in relation to the very existence of automated decision systems 
\cite{ai_now_institute_using_nodate},
\cite{cities_for_digital_rights_declaration_2020},
\cite{council_of_europe_recommendation_2020},\footnote{Cf. "Identifiability of algorithmic decision-making."}
\cite{dataethical_thinkdotank_white_nodate},\footnote{Cf. "Fair communication."}
their purpose, reach, and actual use \cite{ai_now_institute_using_nodate},
the definitions of key concepts and key measures employed (e.g., definitions of automated decision or AI, \cite{ai_now_institute_using_nodate}
of fairness \cite{leslie_understanding_2019}),
the ethical or impact assessment concerning them, 
\cite{ai_now_institute_using_nodate}, \cite{treasury_board_of_canada_secretariat_directive_2019},\footnote{Cf. "Appendix C - Notice."}
their justification \cite{ai_now_institute_using_nodate}, \cite{government_digital_service_and_office_for_artificial_intelligence_uk_guide_2019},\footnote{"Transparency;" see also \cite{schweizerische_eidgenossenschaft_-_der_bundesrat_leitlinien_2020}, cf. "Leitlinie 3."}
the underlying data types and processing methods \cite{council_of_europe_european_2020}, \cite{treasury_board_of_canada_secretariat_directive_2019}, \footnote{Cf. "Appendix C - Notice."}
and their overall quality, often reductively characterized as accuracy \cite{dataethical_thinkdotank_white_nodate}, \footnote{Cf. "Fair communication."}
effectiveness, efficiency \cite{treasury_board_of_canada_secretariat_directive_2019},\footnote{Cf. "Reporting: 6.5.1."}
or ability to support the administration \cite{treasury_board_of_canada_secretariat_directive_2019}.\footnote{"Appendix C - Notice."}  
Post-hoc explanations of the causes of individual specific decision are also invoked \cite{dataethical_thinkdotank_white_nodate}, \cite{government_digital_service_and_office_for_artificial_intelligence_uk_guide_2019},\footnote{Cf. "Transparency."}
\cite{world_economic_forum_ai_2020},\footnote{Cf. "Human in the loop".}
(Government of Canada and Treasury Board Secretariat 2019).\footnote{Cf. "6.2.3."}
Another key form of answerability, namely contestation, is also invoked for automated decisions \cite{automated_decision_systems_task_force_new_2019},\footnote{Cf. "3.2. Incorporate information about ADS specifically…"}
with emphasis on contestation because of the risk of harmful or discriminatory effects \cite{automated_decision_systems_task_force_new_2019},\footnote{Cf. "3.3 Create an internal City process for assessing…"}
often in association with stressing the value of public participation \cite{cities_for_digital_rights_declaration_2020}.

It is acknowledged that it is not reasonable to exact the same level of transparency to be required of all systems \cite{council_of_europe_recommendation_2020},\footnote{Cf. "Levels of transparency."}\cite{government_of_new_zealand_algorithm_2020}.\footnote{Cf. "Transparency."}
Yet, some guidelines characterize transparency to be (what a philosopher would characterize as) a general (pro-tanto) principle, meaning, the highest possible transparency should always be achieved, compatibly with all other overriding (legal and moral) constraints being satisfied \cite{council_of_europe_recommendation_2020}.\footnote{Cf. "Levels of transparency."}
The counterpart of transparency — the actors with entitlement to ask questions and receive truthful information — may vary. As public transparency is at stake here, we only consider counterparts that belong to the broader public, or the people subjected to the authority of public administrations. Most prescriptions consider the individuals involved or affected \cite{dawson_artificial_2020}, \cite{dataethical_thinkdotank_white_nodate},\footnote{Cf. "Transparency." See also See also Schweizerische Eidgenossenschaft - Der Bundesrat. 2020, cf. "Leitlinie 3."}
\cite{government_digital_service_and_office_for_artificial_intelligence_uk_guide_2019},\footnote{Cf. "Ongoing review."} \cite{council_of_europe_recommendation_2020},\footnote{Cf. "Expertise and oversight."}
the public in general \cite{council_of_europe_recommendation_2020}, \footnote{Cf. "Public debate."}
or independent experts \cite{council_of_europe_recommendation_2020},\footnote{Cf. "Expertise and oversight."}
\cite{ai_now_institute_using_nodate}. Even communication by a whistleblower is considered as deserving of encouragement and protection by the laws of the state and the organization of a company \cite{council_of_europe_recommendation_2020}.\footnote{Cf. "Advancement of public benefit."} 
\section{Accountability and auditability}
Section 4 illustrates why transparency is plausibly required by accountability. But is internal transparency also sufficient for accountability? The question here is not whether internal transparency is sufficient for public accountability — our account in section 3 entails that internal transparency is not sufficient for public accountability unless it is paired with some entitlements to transparency and sanctioning. The question is, rather, whether internal transparency can be integrated in a form of public accountability, with some additions. This section explores that possibility.

The fundamental point is that all forms of accountability require, at the minimum, a counterpart outside the accountable organization with some kind of entitlement (legal or de facto) to:
\begin{enumerate}
    \item ask specific questions; 
    \item receive truthful answers.
\end{enumerate} 
The element of sanctionability is also necessary for accountability, but notice that the party entitled to sanction and the party entitled to information access need not be the same.  For example, the right of auditors to receive information may be derived from legal regulation. When the party with an entitlement to access to information (i.e., the auditor) is not given access, or is given non-truthful information, or when the information provided does not fulfill some regulatory standard, the authority to sanction may rest on the judiciary exercising the authority of the law — which is an expression of popular sovereignty.

One can then characterize a distinct form of accountability which includes:
\begin{enumerate}
    \item a party "AU", with special entitlements to transparency i.e., the right to ask certain questions and to receive truthful answers to them;
    \item a party (not necessarily AU), with special entitlements to sanction, and providing the grounds of AU's entitlements to control.
\end{enumerate} 
For ease of exposition, the party designated above as "AU" can be considered an automation auditor, borrowing the terminology from the domain of accounting. Auditors can play:
\begin{enumerate}
    \item an instrumental role in achieving public accountability;
    \item an instrumental role in achieving any of the other ethically desirable outcome for which public transparency is often referred as a means.
\end{enumerate}
Let us analyze both functions in turn. For case (a) we have to assume that the entitlement to transparency of the auditor has a legal basis. This is similar to the case in finance, where the law prescribes that a company's accounts have to be audited and the results of these audits have to be made available to someone — shareholders, tax authorities, an oversight institution etc. The auditors are (often) private companies that have to follow certain rules that are also based on law, and they themselves are controlled by oversight institutions.\footnote{These systems often fail, sometimes spectacularly.} The interesting mechanism here is that while one party has the transparency entitlements (AU), the public, represented by the judiciary, has the sanctioning entitlement, which may be exercised on the auditor, on the audited organization, or both. For example, in the words of the proposed EU Digital Service Act \cite{european_commission_proposal_2020}, auditors "should be accountable, through independent auditing, for their compliance with the obligations laid down by this Regulation and, where relevant, any complementary commitments undertaking pursuant to codes of conduct and crises protocols." In this model, the audited organization is accountable to the auditor (the auditor is entitled to ask questions and receive truthful answers), while both auditors and organizations are sanctionable (by the state). This is a chain-of-accountability model, that amounts to public accountability in an indirect way.

\begin{figure}[ht]
\centering
{\includegraphics[width=1\linewidth]{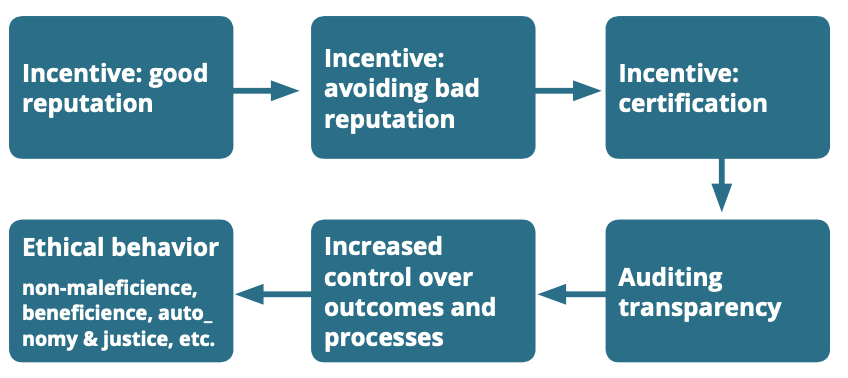}
\label{fig:delegation}}
\hfill
\caption{Internal transparency combined with auditing entitlements generate accountability, which is instrumentally valuable in so far as it incentivizes ethical behavior.}
\end{figure}

Notice that, in the case of the public administration at least, the auditor's entitlement always derives from a legal requirement, so the entitlement to sanction belongs to the public represented as the citizen. In this case, auditing can be considered a means to a structure of accountability that ultimately contributes to democratic self-government. It amounts to a form of indirect control of the public of the confidentially audited party, where control is achieved not directly, but through delegation to specialized parties, each of which owns different entitlements (lawmakers, the judiciary, auditors etc) vis-à-vis the accountable organization. This is a rather different model from the public transparency one sketched in section 3.

This is not the only way in which auditing can contribute to accountability in general. A distinct (moral and legal) entitlement to transparency can originate from contractual agreement. This is more plausible in the use of auditing by private firms, where a manager's decision to be audited need not be legally required. A private company management may rely on auditing instrumentally, i.e., to obtain two goals: an indirect form of control on the processes in the company and (prudentially) a reputation booster. In this case, accountability exists if and only if the two functions are well-aligned, i.e., auditing improves both the level of control over processes (which produces ethically desirable consequences) and the company reputation (which has prudential value). If good reputation and good control are not functionally related, there can be no accountability. For reputation is here the primary currency the public can use to sanction poorly controlled organizations. Even in this case, the auditor is not only a counterpart who just happens to gain information, but one that is entitled to ask specific question and to obtain truthful answers. The entitlement to sanction here can be seen as resting in consumers, who may not be willing to trust a company unless it is audited and certified. 

\section{Conclusion}
In our examination of guidelines, we found that there is little awareness that the different forms of public accountability (direct public accountability and indirect public accountability through auditing) operate by virtue of distinct entitlements. There is also little awareness of the different types of arguments (instrumental vs. non-instrumental) that can be spent in favor of it. Accountability is generally described as a desirable goal, or (more often) as a requirement, but the reason why this is, and what this exactly entails, is often not clarified in them.

The idea of auditing accountability is arguably a lingering background thought, that may explain why so much attention is paid — often under the heading of accountability — to some standard safety and quality control mechanisms that are good business practices but do not alone qualify as accountability unless some entitlement to transparency and sanctioning mechanism also exists. In section 5 we have argued that internal transparency is necessary for responsibility identification, which is a presupposition of accountability. But clearly does not yet entail that internal transparency is sufficient for it. 

Internal transparency can be turned into an independent accountability mechanism only if auditors are entitled to ask questions and receive truthful answers and only if they are, in turn, accountable to the public. Moreover, the problem of many-hands may involve auditors themselves, so clear auditor responsibilities and liabilities must be defined. 

Our analysis has allowed us to distinguish between direct public accountability via public transparency and indirect public accountability via transparency to auditors. In order to do so, we started with a philosophical analysis of the elements of accountability and of delegation from human to computationally driven agents. In particular, we have shown that the key element of accountability, responsibility identification, is clearly addressed in existing guidelines. We have also identified two sets of requirements that are ordinarily associated with transparency, namely, public transparency and internal transparency (or control). These requirements — we have argued — are enablers of accountability: direct public accountability in the former case, and indirect public accountability in the latter. The difference between direct public accountability and indirect public accountability is that in the former, the public itself is expected to control the administration and transparency must be addressed to it. In indirect public accountability, by contrast, the public expresses its right/duty to self-government through its legislators and control is exercised directly by auditors.

We have identified two potentially overlapping normative arguments for public accountability: an instrumental argument, namely that accountable parties are more likely to behave ethically, and a non-instrumental one, namely that under self-government, the governed have a right/duty to control the governors. In relation to the second argument, we have shown that certain forms of automation (those involving imperfect delegation) prevent citizens from exercising this right. From this, a duty to make AI accountable follows. This duty could also be discharged through accountable auditing grounded in law by democratic legislatures. So, in the case of the public sector, auditing (with a legal basis) can also be seen as an indirect form of control by the public. Auditing can also be more generally ethically valuable by virtue of its effects — if and when it incentivizes ethical behavior and other ethically valuable outcome and process improvements. While some guidelines require processes that are technically analogous to auditing, the debate needs more clarity about what the entitlements and liabilities of auditors should be. This is essential for any ethical proposition about auditing being a public accountability device to be valid.
\section{Methodological Appendix}
The authors selected 16 guidelines for examination from 172 in AlgorithmWatch’s AI Ethics Guidelines Global Inventory.\footnote{https://inventory.algorithmwatch.org} The selected guidelines are those directly connected to the public sector, either being written for it, or being an emanation of it.

\begin{table}[h]
\small
\begin{tabular}{|l|l|l|}
\hline
\textbf{} & \textbf{\begin{tabular}[c]{@{}l@{}}Recommendations\\  for the use of \\ AI-based systems\end{tabular}} & \textbf{\begin{tabular}[c]{@{}l@{}}Laws and regulations\\  on the use of \\ AI-based systems\end{tabular}} \\ \hline
\textit{\begin{tabular}[c]{@{}l@{}}In the scope \\ of this study\end{tabular}} & \begin{tabular}[c]{@{}l@{}}Address the public\\  administration\end{tabular} & \begin{tabular}[c]{@{}l@{}}Relate horizontally to \\ AI-based systems / \\ ADM systems\end{tabular} \\ \hline
\textit{\begin{tabular}[c]{@{}l@{}}Outside \\ the scope of \\ this study\end{tabular}} & \begin{tabular}[c]{@{}l@{}}Aimed at all \\ developers and users\end{tabular} & \begin{tabular}[c]{@{}l@{}}Refer to AI-based systems/ \\ ADM systems in a \\ specific sector\end{tabular} \\ \hline
\end{tabular}
\caption{\label{tab:table-name2}Source selection.}
\end{table}
This resulted in the selection of 16 guidelines. The guidelines texts have been coded by one researcher according to a codebook specialized on the contents of AI ethical guidelines, comprising codes for both goals (e.g., avoid discrimination, part of the overall goal of justice) and required actions (e.g., monitoring, as a species of control). This codebook was developed through the analysis of guidelines in previous work of one of the authors, with a combination of inductive and philosophical (a-priori) conceptualization methodology \cite{loi_comparative_2020}.

\begin{acks}
We thank Anna Mätzener and our three anonymous reviewers for their helpful feedback.
\end{acks}

\bibliographystyle{ACM-Reference-Format}
\balance
\bibliography{main}

\end{document}